\pdfoutput=1
\documentclass[11pt]{article}

\usepackage[]{ACL2023}

\usepackage{times}
\usepackage{latexsym}
\usepackage{url}
\usepackage[T1]{fontenc}

\usepackage[utf8]{inputenc}
\usepackage{multirow}
\usepackage{microtype}
\usepackage{booktabs}
\usepackage{inconsolata}

\usepackage{graphicx} 
\title{Enhancing ESG Impact Type Identification through \\ Early Fusion and Multilingual Models}

\author{Hariram Veeramani \\
  Department of Electrical  \\
  and Computer Engineering, \\
  UCLA, USA \\
  \texttt{hariram@ucla.edu} \\\And
  Surendrabikram Thapa \\
  Department of Computer  \\
  Science, Virginia Tech, \\
  Blacksburg, USA \\
  \texttt{sbt@vt.edu} \\\And
  Usman Naseem \\
  College of Science and\\
  Engineering, James Cook \\
  University, Australia \\
  \texttt{usman.naseem@jcu.edu.au} \\}

\begin{document}
\maketitle
\begin{abstract}
In the evolving landscape of Environmental, Social, and Corporate Governance (ESG) impact assessment, the ML-ESG-2 shared task proposes identifying ESG impact types. To address this challenge, we present a comprehensive system leveraging ensemble learning techniques, capitalizing on early and late fusion approaches. Our approach employs four distinct models: mBERT, FlauBERT-base, ALBERT-base-v2, and a Multi-Layer Perceptron (MLP) incorporating Latent Semantic Analysis (LSA) and Term Frequency-Inverse Document Frequency (TF-IDF) features. Through extensive experimentation, we find that our early fusion ensemble approach, featuring the integration of LSA, TF-IDF, mBERT, FlauBERT-base, and ALBERT-base-v2, delivers the best performance. Our system offers a comprehensive ESG impact type identification solution, contributing to the responsible and sustainable decision-making processes vital in today's financial and corporate governance landscape.
\end{abstract}

\section{Introduction}
In the rapidly evolving landscape of finance and corporate governance, Environmental, Social, and Corporate Governance (ESG) considerations have gained unprecedented prominence \cite{linhares-pontes-etal-2022-using}. Investors, stakeholders, and businesses increasingly recognize the importance of ESG factors in decision-making processes, portfolio management, and corporate strategy \cite{li2021esg, gillan2021firms}. The ``E'' in ESG shines a spotlight on environmental sustainability, compelling companies to address pressing issues such as climate change and resource conservation. Recognizing that eco-friendly practices are ethical and financially savvy, businesses are embracing environmental responsibility as a key driver of long-term success \cite{alsayegh2020corporate, newell2022increasing}. Simultaneously, the ``S'' in ESG emphasizes social responsibility, pushing corporations to foster diversity, inclusivity, and community engagement \cite{buallay2019sustainability}. Companies that prioritize these social factors not only enhance their reputation but also attract top talent and build resilience in a socially conscious world. Finally, the ``G'' underscores the importance of sound corporate governance, providing the foundation for trust and stability in financial markets. Investors now realize that transparent decision-making, ethical conduct, and effective risk management are essential elements for ensuring a company's long-term viability \cite{tian-etal-2022-automatic}.


 In this context, the ML-ESG-2 shared task is pivotal in facilitating a deeper understanding of how ESG information can be extracted and utilized from the vast sea of textual data, contributing to the broader endeavor of responsible and sustainable finance. This specific task of ESG impact type identification, where machine learning models classify news articles or sentences as ``Opportunity'', ``Risk'', or ``Cannot Distinguish'', carries profound significance. For investors, this knowledge translates into well-informed decisions with the potential to impact portfolios significantly \cite{amel2018and}. Positive ESG signals can guide investments toward companies poised for sustainable growth while identifying ESG risks enables proactive risk mitigation. In the corporate sphere, grasping the ESG implications of textual data like news articles and annual reports fosters strategic decision-making and risk management. Companies can leverage positive ESG news to enhance their sustainability efforts and attract responsible investors while swiftly addressing negative news to minimize reputational damage and regulatory challenges \cite{goel-etal-2022-tcs}. Furthermore, automated text analysis tools expedite these processes, enabling scalable and efficient ESG information extraction.

As global issues like climate change, social inequality, and ethical governance become increasingly prominent, harnessing ESG information from textual sources is a means to hold corporations accountable. By scrutinizing relevant text like news articles or annual securities reports for their ESG impact, stakeholders can encourage responsible behavior, promote sustainable practices, and ensure that businesses prioritize societal and environmental concerns \cite{kannan-seki-2023-textual}. This task not only empowers decision-makers with the ability to navigate the complexities of ESG information but also contributes to a broader cultural shift towards responsible investment and corporate citizenship. In essence, ESG information extraction from text represents a vital step towards a more sustainable and equitable future, where financial decisions align with environmental and social responsibility goals.

In Natural Language Processing (NLP), text classification is a critical enabler for tasks like ESG impact type identification, bridging the gap between the vast expanse of textual data and actionable insights. NLP techniques have been widely used in the financial domain \cite{jaggi2021text, adhikari2023explainable}. Leveraging NLP techniques, machine learning models can autonomously categorize news articles or sentences into predefined ESG impact types, such as ``Opportunity'', ``Risk'', or ``Cannot Distinguish''. These models harness the power of feature extraction, supervised learning, feature engineering, and even advanced transfer learning from pre-trained language models to accurately assess the implications of textual content on a company's ESG performance. The synergy between NLP and ESG, impact type identification, is instrumental in empowering stakeholders, including investors and corporations, to efficiently navigate the complexities of ESG information, make informed decisions, and contribute to a more sustainable and responsible financial and corporate governance landscape \cite{kang-el-maarouf-2022-finsim4}.

This paper presents the system description of our submission to ML-ESG-2, a shared task hosted by the FinNLP workshop in conjunction with IJCNLP-AACL 2023. In this task, we delve into the critical intersection of Natural Language Processing (NLP) and ESG impact type identification. 

\section{Task Description}

ML-ESG-2 introduces a new task called ESG impact type identification. 

\subsection{Objective}
Given textual data, the objective is to classify it into one of three categories from the ESG aspect: ``Opportunity'', ``Risk'', or ``Cannot Distinguish'' (or ``Positive'', ``Negative'', or ``N/A'' in the Japanese dataset).

\subsection{Dataset}

The ML-ESG-2 task utilizes a carefully curated dataset by \citet{kannan-seki-2023-textual} to support the ESG impact type identification objective. This dataset was curated using Japanese annual securities reports as a primary source of information to extract insights into a company's ESG (Environmental, Social, and Corporate Governance) initiatives. In the context of ESG impact type identification, each news article or sentence is categorized into one of three distinct labels. The label ``Opportunity'' is assigned to articles or sentences that suggest a positive impact or opportunity for the company from an ESG perspective. Conversely, the label ``Risk'' is assigned to articles or sentences that convey a negative impact or potential risk to the company concerning ESG factors. However, in cases where the available content does not provide sufficient clarity to definitively classify the article as either an opportunity or a risk, the label ``Cannot Distinguish'' is used. It is worth noting that in the Japanese dataset, these labels are represented as ``Positive'', ``Negative'', and ``N/A'' to align with the language and terminology commonly used in Japanese ESG contexts. 


\section{System Description}
In this section, we provide a detailed overview of our system architecture and methodology for the ML-ESG-2 shared task on ESG impact type identification. Our approach is founded on the principles of ensemble learning, utilizing both early fusion and late fusion techniques to harness the strengths of multiple language models and feature representations. In order to approach this task, we use four different models.

\begin{figure*}[t]
    \centering
    \includegraphics[width = 0.87\linewidth]{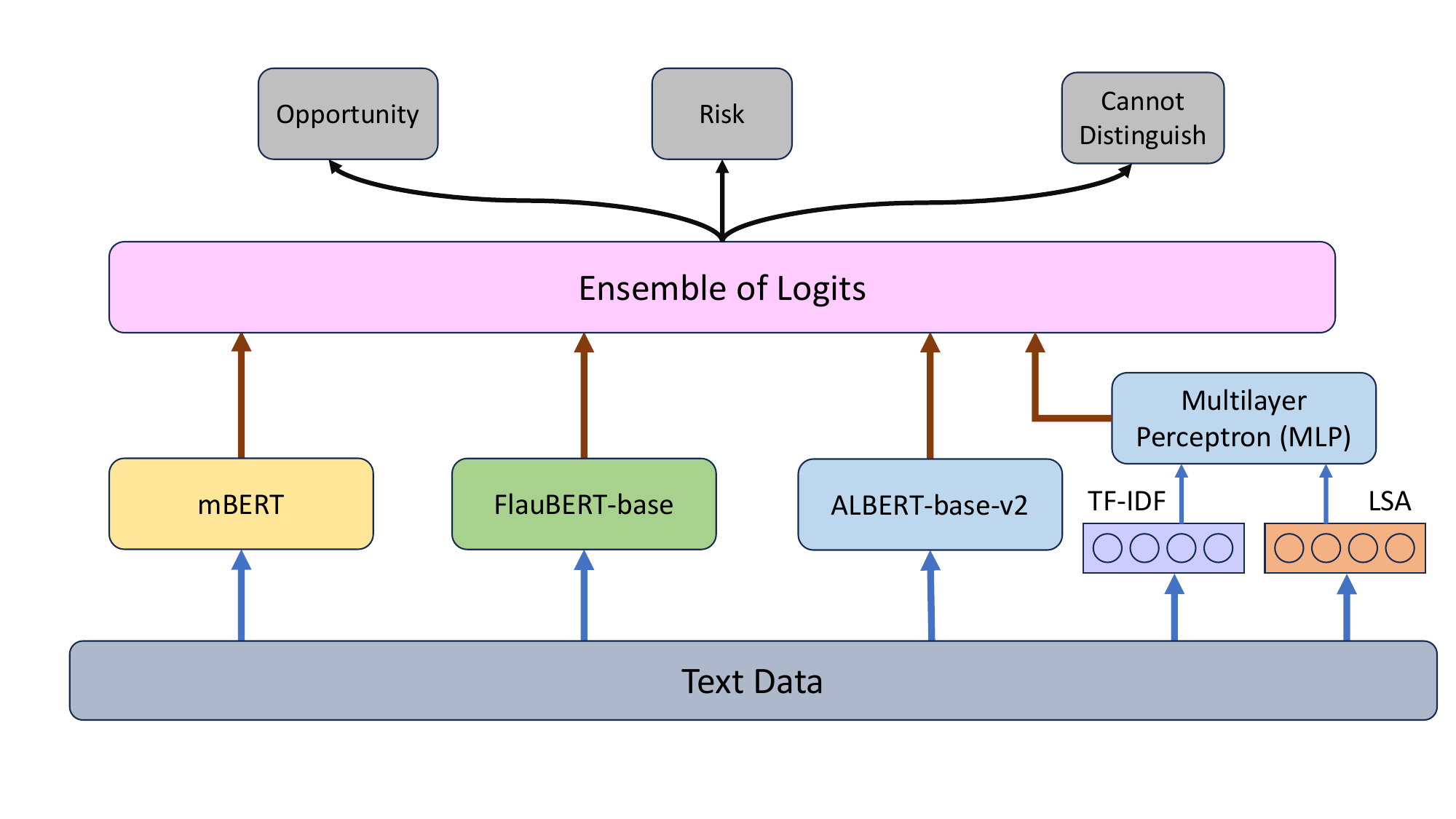}
    \vspace{-0.7cm}
    \caption{Late fusion technique that uses logits from all models to make the final decision.}
    \label{fig:late-fusion}
\end{figure*}

\subsection{Models Used}
In our system, we leverage the predictive capabilities of four different models, each contributing unique strengths to the ESG impact type identification process:

\subsubsection{mBERT (Multilingual BERT)}
To ensure robust multilingual text analysis, we fine-tune mBERT \cite{devlin-etal-2019-bert,wu2020all} on the task-specific dataset. This model adapts to contextual information and linguistic nuances in the news articles, making it a valuable component for handling multilingual content \cite{veeramani-hariram-etal-2023-araieval, veeramani-hariram-etal-2023-Nadi, veeramani-hariram-etal-2023-Quran} effectively. 

\subsubsection{FlauBERT-base (French Language Model)}
Given the presence of French textual content in the task, we incorporate FlauBERT-base \cite{le2020flaubert}, a specialized French language model. This addition ensures accurate analysis and classification of French content, enhancing the overall system's robustness.

\subsubsection{ALBERT-base-v2}
We integrate ALBERT-base-v2 \cite{lan2019albert}, known for its efficient parameterization and performance, to diversify our model components. This model further enhances our system's capability to capture nuanced ESG-related nuances and impact types.

\subsubsection{MLP (Multi-Layer Perceptron)}
Complementing the language models, we employ a Multi-Layer Perceptron (MLP) \cite{murtagh1991multilayer} that incorporates feature representations derived from Latent Semantic Analysis (LSA) \cite{dumais2004latent} and Term Frequency-Inverse Document Frequency (TF-IDF) \cite{bafna2016document, adhikari2021comparative}. These features capture semantic information and term importance within the documents. The MLP facilitates the modeling of intricate feature interactions \cite{veeramani-etal-2023-focal, KeSebErt}, enhancing the depth of our system's analysis. 

\subsection{Ensemble Techniques}
Our ensemble strategy combines both early fusion and late fusion techniques, capitalizing on the collective intelligence of individual models and feature representations.

\subsubsection{Late Fusion Ensemble}
In the late fusion approach, we aggregate predictions from individual models and feature representations at the logits level \cite{imagearg-overview}. This technique effectively combines the diverse outputs generated by mBERT, FlauBERT-base, ALBERT-base-v2, and the MLP with LSA and TF-IDF features. Late fusion ensures comprehensive integration of the predictive capabilities of these components, resulting in a more robust and accurate prediction for ESG impact type identification. Figure \ref{fig:late-fusion} shows the schematic overview of the late fusion ensembling technique we employ.

\begin{figure*}[h]
    \centering
    \includegraphics[width = 0.87\linewidth]{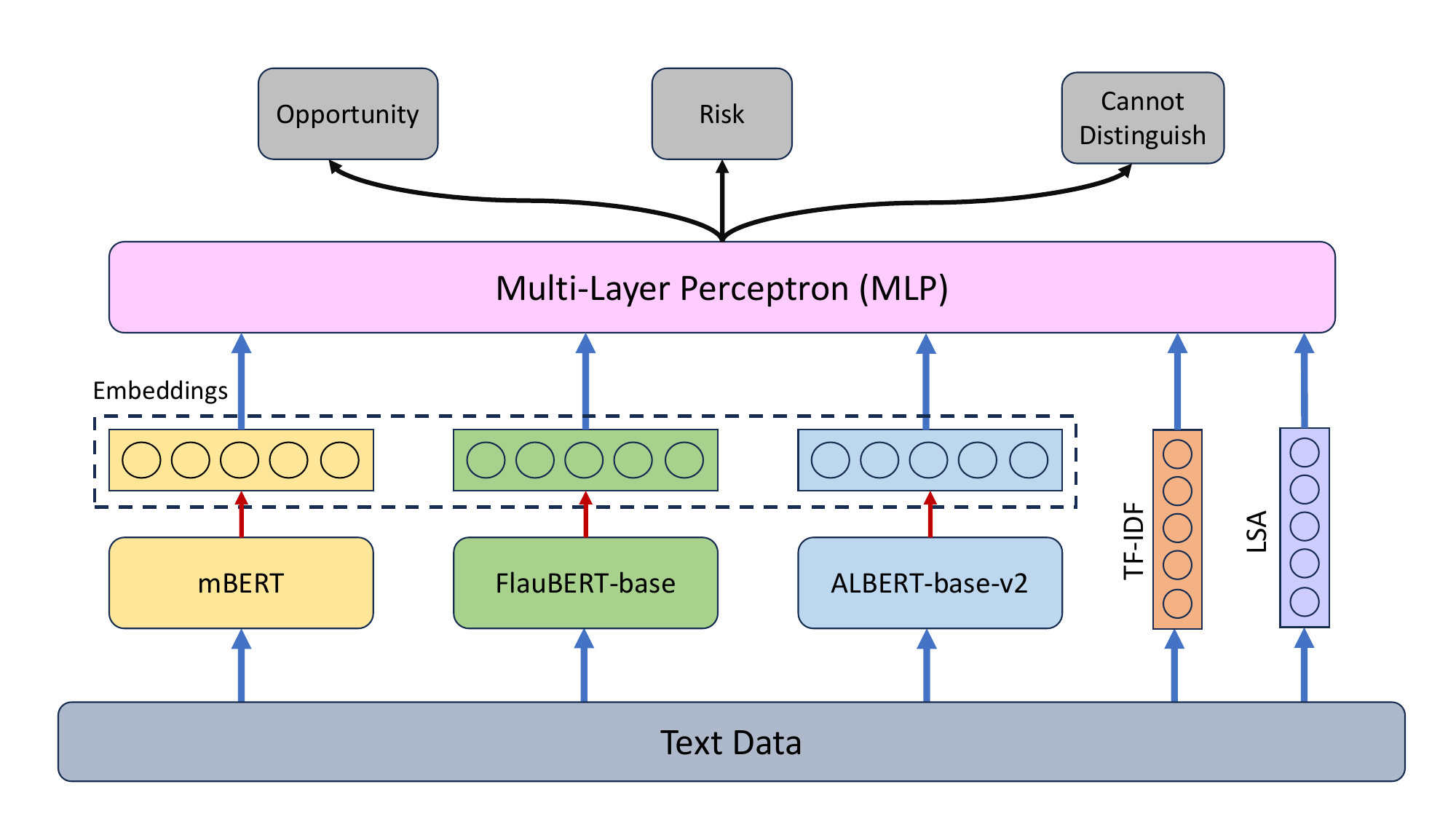}
    \vspace{-0.2cm}
    \caption{Early fusion ensemble takes the different representations and uses MLP for the final classification.}
    \label{fig:early-fusion}
\end{figure*}

\begin{table*}[h]
    \centering
    \begin{tabular}{ccccc}
    \toprule
    Embeddings & Language & Micro-F1 & Macro-F1 & Weighted-F1 \\
    \toprule
         \multirow{4}{*}{\shortstack{FlauBERT + mBERT +\\ALBERT + TF-IDF}} & English & 0.9587 & 0.9128 & 0.9587 \\
         & French & 0.545 & 0.520 & 0.5111\\
         & Japanese & 0.5323 & 0.2933 & 0.4905\\
         & Chinese & 0.403 & 0.1667 & 0.3782\\
         \midrule
         \multirow{4}{*}{\shortstack{FlauBERT + mBERT +\\ALBERT + TF-IDF + LSA}} & English & 0.9633 & 0.918 & 0.9639 \\
         & French & 0.5501 & 0.5292 & 0.5155 \\
         & Japanese & 0.5378 & 0.3043 & 0.4943 \\
         & Chinese & 0.4102 & 0.1728 & 0.3881 \\   
         \bottomrule
    \end{tabular}
    \caption{Ablation study for different combinations of embeddings with MLP layers as the classification layer (early fusion).}
    \label{tab:ablation-early}
\end{table*}

\subsubsection{Early Fusion Ensemble}
In the early fusion approach, we integrate representations from mBERT, FlauBERT-base, ALBERT-base-v2, LSA, and TF-IDF features at an earlier stage, prior to inputting them into the MLP for prediction. Early fusion allows for seamless information integration from multiple sources, empowering the MLP to capture intricate relationships among these elements. This approach ensures the extraction of richer semantic and contextual features, enhancing the system's ability to make nuanced predictions for identifying ESG impact types in news articles. Figure \ref{fig:early-fusion} shows the schematic overview of the early fusion ensembling technique we employ.

\begin{table*}[]
    \centering
    \begin{tabular}{ccccc}
    \toprule
    Models & Language & Micro-F1 & Macro-F1 & Weighted-F1 \\
    \toprule
         \multirow{4}{*}{\shortstack{FlauBERT + mBERT +\\ALBERT}} & English & 0.9403 & 0.8899 &  0.9357 \\
         & French & 0.525 & 0.501 & 0.4933 \\
         & Japanese & 0.5155 & 0.2755 & 0.465 \\
         & Chinese & 0.378 & 0.1346 & 0.3525 \\
         \midrule
         \multirow{4}{*}{\shortstack{FlauBERT + mBERT +\\ALBERT + MLP (TF-IDF)}} & English & 0.945 & 0.8944 & 0.9403 \\
         & French & 0.530 & 0.505 & 0.5022 \\
         & Japanese & 0.520 & 0.280 & 0.475 \\
         & Chinese & 0.384 & 0.141 & 0.3589 \\   
         \midrule
         \multirow{4}{*}{\shortstack{FlauBERT + mBERT +\\ALBERT + MLP (TF-IDF + LSA)}} & English & 0.9495 &  0.8991 & 0.9449 \\
         & French & 0.535 & 0.512 & 0.5066 \\
         & Japanese &  0.5244 & 0.2899 & 0.480 \\
         & Chinese & 0.391 & 0.1474 & 0.3717 \\   
         \bottomrule
    \end{tabular}
    \caption{Ablation study for different combinations of models (late fusion).}
    \label{tab:ablation-late}
\end{table*}

By adopting both early and late fusion ensemble techniques, our system maximizes accuracy and robustness, offering a powerful solution for the task of ESG impact type identification in textual data.

\section{Results}
In the ML-ESG-2 shared task for ESG impact type identification, we evaluated the performance of our system across multiple languages using different ensemble configurations and fusion techniques. The micro F1, macro F1, and weighted F1 scores offer a comprehensive assessment of our system's effectiveness in capturing nuances within different linguistic contexts.

\begin{table}[h]
    \centering
    \small
    \begin{tabular}{cccc}
    \toprule
         Team &	Micro-F1 &	Macro-F1 &	Weighted-F1 \\
         \toprule
        AnakItik &	0.9817 &	0.9548 &	0.9810 \\
        BrothFink &	0.9771 &	0.9445 &	0.9765 \\
        NeverCareU &	0.9633 &	0.9227 &	0.9648 \\
        \textbf{FinNLU} &	\underline{0.9633} &	\underline{0.9180} & \underline{0.9639} \\
        231 &	0.9633 &	0.9127 &	0.9627 \\
        SPEvFT & 0.9587 &	0.9118 &	0.9602 \\
        LIPI &	0.9312 &	0.8335 &	0.9294 \\
        HHU &	0.9174 &	0.8098 &	0.9174 \\
        \bottomrule
    \end{tabular}
    \caption{Leaderboard for ESG identification in English language}
    \label{tab:leaderboard}
\end{table}

Our system secured notable rankings in the shared task, claiming the fourth position for English, the third position for both Japanese and Chinese, and the fifth position for the French language. We employed various ensemble configurations, including late fusion models combining FLAUBERT, mBERT, and ALBERT, incorporating TF-IDF and Latent Semantic Analysis (LSA) features. The results demonstrate significant improvements when transitioning to early fusion techniques, particularly when combining TF-IDF, LSA, and language model representations, as shown in Table \ref{tab:ablation-early}. The late fusion techniques, adding more information with multiple models, seemed to help, as shown in Table \ref{tab:ablation-late}.

In the English language, our best early fusion ensemble achieved a micro F1 score of 0.9633, a macro F1 score of 0.918, and a weighted F1 score of 0.9639, showcasing its proficiency in ESG impact type identification. Similarly, for other languages, the early fusion ensemble, which integrates TF-IDF and LSA features with language models, achieved the highest performance among all early fusion and late fusion techniques.

While our ensemble system demonstrates commendable performance in English (as shown in Table \ref{tab:leaderboard}), it is evident that its effectiveness diminishes when applied to other languages such as French, Japanese, and Chinese. This discrepancy underscores the importance of adapting our approach to address language-specific nuances and complexities. To remedy this, future research efforts may focus on language-specific model fine-tuning, dataset augmentation with more diverse linguistic content, and the development of specialized language models tailored to the unique challenges posed by each language. Additionally, incorporating more extensive language-specific features and linguistic resources into our ensemble configurations could further enhance the system's adaptability and robustness across diverse linguistic contexts. These measures hold the potential to improve our system's performance and make it a valuable tool for multilingual ESG impact type identification.

\section{Conclusion}
In the rapidly evolving finance and corporate governance landscape, the ML-ESG-2 shared task has presented a unique opportunity to explore ESG impact type identification. Our comprehensive system leverages the collective intelligence of multiple language models and feature representations through advanced ensemble learning techniques, resulting in high performance. While our system excels in English, it faces hurdles in accurately identifying ESG impact types in Japanese, Chinese, and French languages. These challenges underscore the importance of further research and fine-tuning to adapt our approach to diverse linguistic contexts.

\section*{Limitations}
While our ensemble learning approach has demonstrated strong performance, we acknowledge certain limitations in our study. The system's performance variation across languages, with lower accuracy for Japanese, Chinese, and French texts, highlights the need for further language-specific model development and fine-tuning. Additionally, the reliance on single-source textual data alone may not capture all relevant context and nuances, which may affect the system's ability to identify ESG impact types accurately in certain cases. Furthermore, the generalization of our approach to other ESG-related tasks and domains may require additional adaptation and validation. We also recognize the evolving nature of language models and the need for continuous updates and refinements to maintain their effectiveness.

\section*{Ethics Statement}
Throughout our research, we recognize that ethical considerations, particularly those related to bias mitigation and fairness in model predictions, represent ongoing and critical areas for improvement in the development of responsible AI systems. It is important to note that, in this study, we have not undertaken specific measures to actively mitigate bias within our models or predictions.


\begin{thebibliography}{27}
\expandafter\ifx\csname natexlab\endcsname\relax\def\natexlab#1{#1}\fi

\bibitem[{Adhikari et~al.(2023)Adhikari, Thapa, Naseem, Lu, Bharathy, and Prasad}]{adhikari2023explainable}
Surabhi Adhikari, Surendrabikram Thapa, Usman Naseem, Hai~Ya Lu, Gnana Bharathy, and Mukesh Prasad. 2023.
\newblock Explainable hybrid word representations for sentiment analysis of financial news.
\newblock \emph{Neural Networks}, 164:115--123.

\bibitem[{Adhikari et~al.(2021)Adhikari, Thapa, Singh, Huo, Bharathy, and Prasad}]{adhikari2021comparative}
Surabhi Adhikari, Surendrabikram Thapa, Priyanka Singh, Huan Huo, Gnana Bharathy, and Mukesh Prasad. 2021.
\newblock A comparative study of machine learning and nlp techniques for uses of stop words by patients in diagnosis of alzheimer's disease.
\newblock In \emph{2021 International Joint Conference on Neural Networks (IJCNN)}, pages 1--8. IEEE.

\bibitem[{Alsayegh et~al.(2020)Alsayegh, Abdul~Rahman, and Homayoun}]{alsayegh2020corporate}
Maha~Faisal Alsayegh, Rashidah Abdul~Rahman, and Saeid Homayoun. 2020.
\newblock Corporate economic, environmental, and social sustainability performance transformation through esg disclosure.
\newblock \emph{Sustainability}, 12(9):3910.

\bibitem[{Amel-Zadeh and Serafeim(2018)}]{amel2018and}
Amir Amel-Zadeh and George Serafeim. 2018.
\newblock Why and how investors use esg information: Evidence from a global survey.
\newblock \emph{Financial analysts journal}, 74(3):87--103.

\bibitem[{Bafna et~al.(2016)Bafna, Pramod, and Vaidya}]{bafna2016document}
Prafulla Bafna, Dhanya Pramod, and Anagha Vaidya. 2016.
\newblock Document clustering: Tf-idf approach.
\newblock In \emph{2016 International Conference on Electrical, Electronics, and Optimization Techniques (ICEEOT)}, pages 61--66. IEEE.

\bibitem[{Buallay(2019)}]{buallay2019sustainability}
Amina Buallay. 2019.
\newblock Is sustainability reporting (esg) associated with performance? evidence from the european banking sector.
\newblock \emph{Management of Environmental Quality: An International Journal}, 30(1):98--115.

\bibitem[{Devlin et~al.(2019)Devlin, Chang, Lee, and Toutanova}]{devlin-etal-2019-bert}
Jacob Devlin, Ming-Wei Chang, Kenton Lee, and Kristina Toutanova. 2019.
\newblock \href {https://doi.org/10.18653/v1/N19-1423} {{BERT}: Pre-training of deep bidirectional transformers for language understanding}.
\newblock In \emph{Proceedings of the 2019 Conference of the North {A}merican Chapter of the Association for Computational Linguistics: Human Language Technologies, Volume 1 (Long and Short Papers)}, pages 4171--4186, Minneapolis, Minnesota. Association for Computational Linguistics.

\bibitem[{Dumais(2004)}]{dumais2004latent}
Susan~T Dumais. 2004.
\newblock Latent semantic analysis.
\newblock \emph{Annual Review of Information Science and Technology (ARIST)}, 38:189--230.

\bibitem[{Gillan et~al.(2021)Gillan, Koch, and Starks}]{gillan2021firms}
Stuart~L Gillan, Andrew Koch, and Laura~T Starks. 2021.
\newblock Firms and social responsibility: A review of esg and csr research in corporate finance.
\newblock \emph{Journal of Corporate Finance}, 66:101889.

\bibitem[{Goel et~al.(2022)Goel, Chauhan, Sangwan, Verma, Dasgupta, and Dey}]{goel-etal-2022-tcs}
Tushar Goel, Vipul Chauhan, Suyash Sangwan, Ishan Verma, Tirthankar Dasgupta, and Lipika Dey. 2022.
\newblock \href {https://doi.org/10.18653/v1/2022.finnlp-1.32} {{TCS} {WITM} 2022@{F}in{S}im4-{ESG}: Augmenting {BERT} with linguistic and semantic features for {ESG} data classification}.
\newblock In \emph{Proceedings of the Fourth Workshop on Financial Technology and Natural Language Processing (FinNLP)}, pages 235--242, Abu Dhabi, United Arab Emirates (Hybrid). Association for Computational Linguistics.

\bibitem[{Jaggi et~al.(2021)Jaggi, Mandal, Narang, Naseem, and Khushi}]{jaggi2021text}
Mukul Jaggi, Priyanka Mandal, Shreya Narang, Usman Naseem, and Matloob Khushi. 2021.
\newblock Text mining of stocktwits data for predicting stock prices.
\newblock \emph{Applied System Innovation}, 4(1):13.

\bibitem[{Kanagasabai et~al.(2023)Kanagasabai, Rajamanickam, Veeramani, Westerski, and Jung~Jae}]{imagearg-overview}
Rajaraman Kanagasabai, Saravanan Rajamanickam, Hariram Veeramani, Adam Westerski, and Kim Jung~Jae. 2023.
\newblock {Semantists at ImageArg-2023: Exploring Cross-modal Contrastive and Ensemble Models for Multimodal Stance and Persuasiveness Classification}.
\newblock In \emph{Proceedings of the 10th Workshop on Argument Mining}, Online and in Singapore. Association for Computational Linguistics.

\bibitem[{Kang and El~Maarouf(2022)}]{kang-el-maarouf-2022-finsim4}
Juyeon Kang and Ismail El~Maarouf. 2022.
\newblock \href {https://doi.org/10.18653/v1/2022.finnlp-1.28} {{F}in{S}im4-{ESG} shared task: Learning semantic similarities for the financial domain. extended edition to {ESG} insights}.
\newblock In \emph{Proceedings of the Fourth Workshop on Financial Technology and Natural Language Processing (FinNLP)}, pages 211--217, Abu Dhabi, United Arab Emirates (Hybrid). Association for Computational Linguistics.

\bibitem[{Kannan and Seki(2023)}]{kannan-seki-2023-textual}
Naoki Kannan and Yohei Seki. 2023.
\newblock \href {https://aclanthology.org/2023.finnlp-1.4} {Textual evidence extraction for {ESG} scores}.
\newblock In \emph{Proceedings of the Fifth Workshop on Financial Technology and Natural Language Processing and the Second Multimodal AI For Financial Forecasting}, pages 45--54, Macao. -.

\bibitem[{Lan et~al.(2019)Lan, Chen, Goodman, Gimpel, Sharma, and Soricut}]{lan2019albert}
Zhenzhong Lan, Mingda Chen, Sebastian Goodman, Kevin Gimpel, Piyush Sharma, and Radu Soricut. 2019.
\newblock Albert: A lite bert for self-supervised learning of language representations.
\newblock In \emph{International Conference on Learning Representations}.

\bibitem[{Le et~al.(2020)Le, Vial, Frej, Segonne, Coavoux, Lecouteux, Allauzen, Crabb{\'e}, Besacier, and Schwab}]{le2020flaubert}
Hang Le, Lo{\"\i}c Vial, Jibril Frej, Vincent Segonne, Maximin Coavoux, Benjamin Lecouteux, Alexandre Allauzen, Benoit Crabb{\'e}, Laurent Besacier, and Didier Schwab. 2020.
\newblock Flaubert: Unsupervised language model pre-training for french.
\newblock In \emph{Proceedings of the Twelfth Language Resources and Evaluation Conference}, pages 2479--2490.

\bibitem[{Li et~al.(2021)Li, Wang, Sueyoshi, and Wang}]{li2021esg}
Ting-Ting Li, Kai Wang, Toshiyuki Sueyoshi, and Derek~D Wang. 2021.
\newblock Esg: Research progress and future prospects.
\newblock \emph{Sustainability}, 13(21):11663.

\bibitem[{Linhares~Pontes et~al.(2022)Linhares~Pontes, Ben~Jannet, Moreno, and Doucet}]{linhares-pontes-etal-2022-using}
Elvys Linhares~Pontes, Mohamed Ben~Jannet, Jose~G. Moreno, and Antoine Doucet. 2022.
\newblock \href {https://doi.org/10.18653/v1/2022.finnlp-1.29} {Using contextual sentence analysis models to recognize {ESG} concepts}.
\newblock In \emph{Proceedings of the Fourth Workshop on Financial Technology and Natural Language Processing (FinNLP)}, pages 218--223, Abu Dhabi, United Arab Emirates (Hybrid). Association for Computational Linguistics.

\bibitem[{Murtagh(1991)}]{murtagh1991multilayer}
Fionn Murtagh. 1991.
\newblock Multilayer perceptrons for classification and regression.
\newblock \emph{Neurocomputing}, 2(5-6):183--197.

\bibitem[{Newell and Marzuki(2022)}]{newell2022increasing}
Graeme Newell and Muhammad~Jufri Marzuki. 2022.
\newblock The increasing importance of environmental sustainability in global real estate investment markets.
\newblock \emph{Journal of Property Investment \& Finance}, 40(4):411--429.

\bibitem[{Tian et~al.(2022)Tian, Zhang, and Chen}]{tian-etal-2022-automatic}
Ke~Tian, Zepeng Zhang, and Hua Chen. 2022.
\newblock \href {https://doi.org/10.18653/v1/2022.finnlp-1.30} {{Automatic Term and Sentence Classification Via Augmented Term and Pre-trained language model in ESG Taxonomy texts}}.
\newblock In \emph{Proceedings of the Fourth Workshop on Financial Technology and Natural Language Processing (FinNLP)}, pages 224--227, Abu Dhabi, United Arab Emirates (Hybrid). Association for Computational Linguistics.

\bibitem[{Veeramani et~al.(2023{\natexlab{a}})Veeramani, Thapa, and Naseem}]{veeramani-etal-2023-focal}
Hariram Veeramani, Surendrabikram Thapa, and Usman Naseem. 2023{\natexlab{a}}.
\newblock {Automated Citation Function Classification and Context Extraction in Astrophysics: Leveraging Paraphrasing and Question Answering}.
\newblock In \emph{Proceedings of the second Workshop on Information Extraction from Scientific Publications}, Online. Association for Computational Linguistics.

\bibitem[{Veeramani et~al.(2023{\natexlab{b}})Veeramani, Thapa, and Naseem}]{veeramani-hariram-etal-2023-Nadi}
Hariram Veeramani, Surendrabikram Thapa, and Usman Naseem. 2023{\natexlab{b}}.
\newblock {DialectNLU at NADI 2023 Shared Task: Transformer Based MultiTask Approach Jointly Integrating Dialect and Machine Translation Tasks}.
\newblock In \emph{Proceedings of The First Arabic Natural Language Processing Conference (ArabicNLP 2023)}, Singapore. Association for Computational Linguistics.

\bibitem[{Veeramani et~al.(2023{\natexlab{c}})Veeramani, Thapa, and Naseem}]{veeramani-hariram-etal-2023-araieval}
Hariram Veeramani, Surendrabikram Thapa, and Usman Naseem. 2023{\natexlab{c}}.
\newblock {KnowTellConvince at ArAIEval 2023: Disinformation and Persuasion Detection using Similar and Contrastive Representation Alignment}.
\newblock In \emph{Proceedings of The First Arabic Natural Language Processing Conference (ArabicNLP 2023)}, Singapore. Association for Computational Linguistics.

\bibitem[{Veeramani et~al.(2023{\natexlab{d}})Veeramani, Thapa, and Naseem}]{veeramani-hariram-etal-2023-Quran}
Hariram Veeramani, Surendrabikram Thapa, and Usman Naseem. 2023{\natexlab{d}}.
\newblock {LowResContextQA at Qur'an QA 2023 Shared Task: Temporal and Sequential Representation Augmented Question Answering Span Detection}.
\newblock In \emph{Proceedings of The First Arabic Natural Language Processing Conference (ArabicNLP 2023)}, Singapore. Association for Computational Linguistics.

\bibitem[{Veeramani et~al.(2023{\natexlab{e}})Veeramani, Thapa, and Naseem}]{KeSebErt}
Hariram Veeramani, Surendrabikram Thapa, and Usman Naseem. 2023{\natexlab{e}}.
\newblock {Temporally Dynamic Session-Keyword Aware Sequential Recommendation system}.
\newblock In \emph{2023 International Conference on Data Mining Workshops (ICDMW)}.

\bibitem[{Wu and Dredze(2020)}]{wu2020all}
Shijie Wu and Mark Dredze. 2020.
\newblock Are all languages created equal in multilingual bert?
\newblock In \emph{Proceedings of the 5th Workshop on Representation Learning for NLP}, pages 120--130.

\end{thebibliography}
\bibliographystyle{acl_natbib}




\end{document}